\documentclass[lettersize,journal]{IEEEtran}
\usepackage{amsmath,amsfonts}
\usepackage{algorithmic}
\usepackage{algorithm}
\usepackage{array}
\usepackage[caption=false,font=normalsize,labelfont=sf,textfont=sf]{subfig}
\usepackage{textcomp}
\usepackage{stfloats}
\usepackage{url}
\usepackage{verbatim}
\usepackage{graphicx}
\usepackage{cite}
\usepackage{float}
\usepackage{color}
\usepackage{pifont}
\usepackage{caption2}
\usepackage{tikz}
\usepackage{xcolor}
\usepackage{bm}
\usepackage[implicit=false]{hyperref}

\hypersetup{hidelinks,
	colorlinks=true,
	allcolors=black,
	pdfstartview=Fit,
	breaklinks=true}

\definecolor{lime}{HTML}{A6CE39}
\DeclareRobustCommand{\orcidicon}{
\begin{tikzpicture}
\draw[lime, fill=lime] (0,0)
circle[radius=0.16]
node[white]{{\fontfamily{qag}\selectfont \tiny \.{I}D}};
\end{tikzpicture}
\hspace{-2mm}
}
\foreach \x in {A, ..., Z}{%
\expandafter\xdef\csname orcid\x\endcsname{\noexpand\href{https://orcid.org/\csname orcidauthor\x\endcsname}{\noexpand\orcidicon}}
}

\begin{document}

\title{Lightweight Self-Knowledge Distillation with  Multi-source Information Fusion}

\author{Xucong Wang\orcidA{}, Pengchao Han, Lei Guo

\thanks{Xucong Wang is with the School of Computer and Communication Engineering, Northeastern University at Qinhuangdao, Qinhuangdao 066000, China (e-mail:XucongWang111@outlook.com)}
\thanks{Pengchao Han (Corresponding author) is with the School of Science and Engineering, The Chinese University of Hong Kong, Shenzhen, Shenzhen 518172, China (e-mail: hanpengchao@cuhk.edu.cn)}
\thanks{ Lei Guo is with the School of Communication and Information Engineering, 
		Chongqing University of Posts and Telecommunications, Chongqing 400065, China  (e-mail: guolei@cqupt.edu.cn). }
  }

\markboth{IEEE TRANSACTIONS ON NEURAL NETWORKS AND LEARNING SYSTEMS}%
{Shell \MakeLowercase{\textit{et al.}}: A Sample Article Using IEEEtran.cls for IEEE Journals}


\maketitle

\begin{abstract}
Knowledge Distillation (KD) is a powerful technique for transferring knowledge between neural network models, where a pre-trained teacher model is used to facilitate the training of the target student model. However, the availability of a suitable teacher model is not always guaranteed. To address this challenge, Self-Knowledge Distillation (SKD) attempts to construct a teacher model from itself. Existing SKD methods add Auxiliary Classifiers (AC) to intermediate layers of the model or use the history models and models with different input data within the same class. However, these methods are computationally expensive and only capture time-wise and class-wise features of data. In this paper, we propose a lightweight SKD framework that utilizes multi-source information to construct a more informative teacher. Specifically, we introduce a Distillation with Reverse Guidance (DRG) method that considers different levels of information extracted by the model, including edge, shape, and detail of the input data, to construct a more informative teacher. Additionally, we design a Distillation with Shape-wise Regularization (DSR) method that ensures a consistent shape of ranked model output for all data. We validate the performance of the proposed DRG, DSR, and their combination through comprehensive experiments on various datasets and models. Our results demonstrate the superiority of the proposed methods over baselines (up to 2.87\%) and state-of-the-art SKD methods (up to 1.15\%), while being computationally efficient and robust. The code is available at \url{https://github.com/xucong-parsifal/LightSKD}.
\end{abstract}


\section{Introduction}
\IEEEPARstart{K}{nowledge} distillation (KD)\cite{hinton2015distilling} is powerful for transferring knowledge between neural network models, enabling model compression, performance improvement, and interpretation. In the vanilla KD framework, a pre-trained larger model acts as the teacher to facilitate training of a smaller model, i.e., the student, for efficient feature characterization of the training dataset. However, the availability of a suitable teacher model is not always guaranteed. KD can be utilized for characterizing the features of the training dataset to improve the performance of a single model, known as self-knowledge distillation (SKD) \cite{furlanello2018born, zhang2019your, ji2021refine, li2022distilling}. 
In SKD, a neural network model acts as its own teacher by utilizing the knowledge extracted from itself to guide its own model training, resulting in improved model accuracy.

\begin{figure}[t!]
\begin{center}
\centerline{\includegraphics[width=\columnwidth]{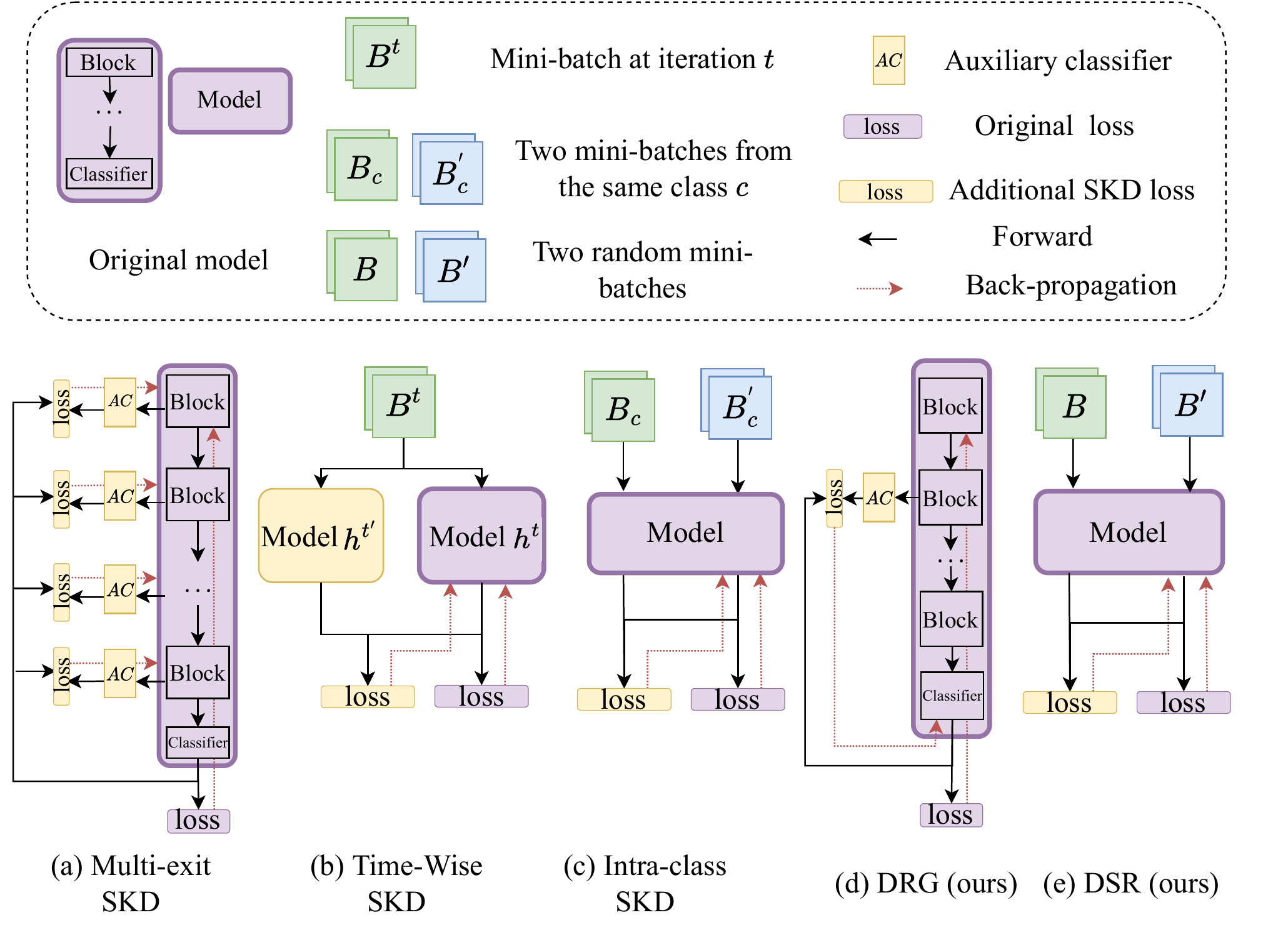}}
\caption{Overview of existing SKD methods, i.e., multi-exit SKD, TW-SKD, and IC-SKD, and our methods, i.e., DRG and DSR.}
\label{fig:comparison}
\end{center}
\end{figure}

To support SKD, existing works have explored various methods for extracting useful knowledge from a model itself, as shown in Fig.~\ref{fig:comparison}.
In general, a neural network model can be divided into several blocks. Each block may contain one or multiple layers in the model. Based on this model architecture, a popular SKD approach named Multi-exit SKD \cite{zhang2019your,ji2021refine,li2022distilling, phuong2019distillation}  is to re-train the early layers (also known as shallow layers) of the model under the guidance of counterpart's outputs or the model's own final output, as shown in  Fig.~\ref{fig:comparison} (a). For example, Be Your Own Teacher (BYOT) \cite{zhang2019your} adds an Auxiliary Classifier (AC) to each block of the model. It uses the knowledge extracted from the final output of the model to train the ACs and update corresponding blocks. 
Multi-exit SKD helps to ensure that all blocks in the model fully learn the features of the training dataset. However, it introduces a high computational overhead for training the additional ACs. For instance, it takes over 5 hours to train BYOT on the CIFAR100 dataset using the ResNet-101 model, compared with about 3.48 hours for training the original model.

Existing SKD methods in the literature with less computational cost use regularization methods that leverage information from history models (i.e., time-wise SKD (TW-SKD)) \cite{yang2019snapshot, kim2021self, shen2022self, yuan2020revisiting, szegedy2016rethinking, kim2022ai, goodfellow2020generative},
as shown in Fig.~\ref{fig:comparison} (b)  and the predictions from the same class of input data  (i.e., intra-class SKD (IC-SKD)) \cite{yun2020regularizing,xu2019data} as shown in Fig.~\ref{fig:comparison} (c). 
TW-SKD methods, such as self-Distillation from the Last mini-Batch (DLB) \cite{shen2022self}, leverage the idea that a ``poor'' teacher that has a low model accuracy may provide useful knowledge compared to a well-trained teacher \cite{yuan2020revisiting} and use historical models as the ``poor'' teacher. 
However, the output of the historical model can only provide limited highly abstracted and inexplicable knowledge on account that model at different training stages learns different levels of features of the input data. 
IC-SKD aims to learn a more generalized model output probability distribution for each class of data by minimizing the distance between the model outputs of different data that belong to the same class. However, IC-SKD overlooks the similarity of inter-class model output probability distributions, which can result in limited model performance and overfitting.

In this paper, we aim to answer the following key question: \textit{How to design SKD to capture more complete features of input data with relatively low computation cost such that to promote model performance?}

We answer the above question by developing a novel informative teacher and learning a consistent shape of the model outputs of all data regardless of their belonging classes. Note that, the informative teacher does not mean that the teacher has a high model accuracy. 
Specifically, preliminary experiments suggest that different layers in a neural network can extract different levels of features for the input data. Typically, shallower layers can capture more shape and edge information while deeper layers can learn more semantic information. This motivates us to construct a teacher by utilizing the feature extracted from the shallow layers to guide the training of the whole model. 
Therefore, we propose Distillation with Reverse Guidance (DRG). DRG employs an AC for a shallow layer and uses the output of the AC to facilitate the student, i.e., the whole model, in learning the shape and edge information from the shallow layer. Thus, the model can simultaneously capture both structural and detailed features of input data, leading to improved model performance. DRG overcomes the high computation cost of BYOT and is able to extract more informative information than TW-SKD.

Furthermore, to learn a consistent shape of the model outputs for all data, we propose Distillation with Shape-wise Regularization (DSR) that aims to explore the shape of inter-class similarity. Different from vanilla KD, where the student mimics the model output distribution of the teacher, and IC-SKD, which focuses on intra-class similarity, DSR learns a consistently ranked model output shape of all data. Our experimental results show that DSR enlarges the decision boundary among classes, contributing to increased model performance.

Our contribution can be summarized as follows:
\begin{itemize}
\item We design a lightweight SKD framework with multi-source information fusion to improve model performance at a low computation cost.
\item We proposed the DRG method that constructs an informative teacher utilizing the output of a shallow layer to facilitate the model simultaneously learning the structural and detailed features of data.
\item We propose the DSR method to stimulate the model learning a consistent ranked output shape of all data regardless of their belonging classes.
\item We evaluate the performance of proposed DRG and DSR methods and their combination over a variety of datasets and models. Notably, our proposed methods outperform the baseline methods by an average of 2\% and the state-of-the-art (SOTA) up to 1.15\%.  
\item We analyze the rationality behind DRG and DSR through experiments and show their superiority in capturing more complete features of data than baselines and enlarging the decision boundary.
\end{itemize}

The remainder of this paper is organized as follows. Section II reviews the related works of KD and SKD. We present preliminaries for the SKD problem in Section III and propose our DRG and DSR methods in Section IV. Sections V and VI demonstrate the experimental results and ablation study, respectively. Section VII discusses the rationality behind DRG and DSR. Finally, Section VIII concludes our paper. 

\begin{figure*}[h]
\begin{center}
\centerline{\includegraphics[width=2\columnwidth]{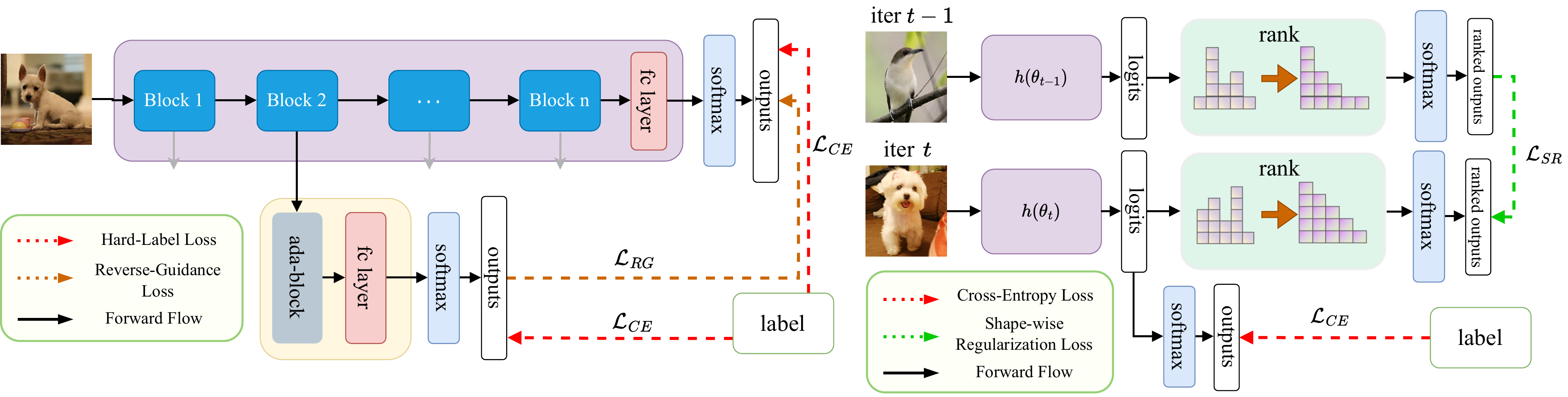}}
\caption{Illustrations of proposed methods. Left: \textbf{DRG}, where an AC is added to the output of a shallow layer to construct a ``poor'' teacher to guide the whole model training. Right:  \textbf{DSR}, where model outputs are ranked to form a inter-class regularization. }
\label{fig:methods}
\end{center}
\end{figure*}

\section{Related Work}
\textbf{Knowledge distillation.} Vanilla KD employs a teacher-student framework to facilitate the student learning from the model output logits of the teacher \cite{hinton2015distilling}\cite{kim2018paraphrasing}. 
A unique parameter in KD is the temperature in the softmax function over the teacher's model output logit, by tuning which, the student can benefit more from the teacher with improved model performance
\cite{li2022curriculum}\cite{li2022asymmetric}.
An improved KD method is feature-based distillation, where the student learns the teacher's intermediate feature \cite{romero2014fitnets}\cite{heo2019comprehensive}\cite{ji2021refine}. 
Works in the literature also have focused on the privacy issues of KD, such as data-free KD that preserves an inaccessible training dataset of the teacher for the student \cite{chen2019data, binici2022robust, zhao2022decoupled}, private model compression \cite{wang2019private}, and undistillable model that prevents a student from learning from the model through KD \cite{ma2021undistillable,kundu2021analyzing,jandial2022distilling}.

\textbf{Self-knowledge distillation.} 
The first SKD work can date back to Born Again Neural Networks (BAN)\cite{furlanello2018born}. BAN employs a serial-distillation mechanism, namely asking teachers to guide students with the same architecture which would later be asked to guide other sub-students. The average of all students' outputs are considered as the final outputs. BYOT  et. al\cite{zhang2019your,ji2021refine,li2022distilling} developed a multi-exit architecture for a neural network. The final output of the network is utilized to update the shallow layers. However, BYOT exerts a high computation cost due to the training of ACs for each exit of the model.

In addition, works in the literature also achieve SKD well by designing a much more delicate regularization to improve model performance. There are three categories of regularization, i.e., TW-SKD, IC-SKD, and SKD with Label Smoothing. TW-SKD uses the model in the history as the teacher to regularize the current model. Specifically, Snapshot distillation (SS-KD) \cite{yang2019snapshot} randomly chooses a model from previous iterations. Progressive refinement knowledge distillation (PS-KD) \cite{kim2021self} and DLB \cite{shen2022self} regard the model in the last epoch as poor-teacher.  
For IC-SKD, the class-wise SKD (CS-KD) \cite{yun2020regularizing} uses two batched of data samples from the same class and minimizes the output discrepancy between the two batches. Data-Distortion Guided Self-Distillation (DDGSD) \cite{xu2019data} exerts different pre-processing techniques on the same batch and minimizes their model output difference. 

Another way to improve the performance of SKD is label-smoothing. The essence of many label-smoothing works lies in the utility of self-teaching, and they can be viewed as special cases of SKD. Label-Smoothing Regularization (LSR) \cite{szegedy2016rethinking} introduces a method where the ground truth distribution is combined with a uniform distribution to create a virtual teacher with random accuracy \cite{yuan2020revisiting}.  Delving Deep into Label Smoothing (OLS) \cite{zhang2021delving} proposes a more reasonable smoothing technique that constructs labels based on the integrated output information from previous epochs.
Inspired by the widespread Zipf's law, Efficient one pass self-distillation with ZipF's Label Smoothing (ZF-LS) \cite{liang2022efficient} seeks to discover the conformality of ranked outputs and introduces a novel counterpart discrepancy loss, minimizing with Zipf's distribution based on self-knowledge.
Motivated by ZF-LS, it is promising to achieve consistent model outputs of all data by using ranked outputs from the last iteration as softened targets, which can be seen as a specific form of label smoothing. 
For SKD with label-smoothing, Teacher-free knowledge distillation (TF-KD) \cite{yuan2020revisiting}, has discovered the entity of Label Smoothing Regularization (LSR) \cite{szegedy2016rethinking} to generate high-accuracy virtual teacher. Adversarial Learning and Implicit regularization for self-Knowledge Distillation  (AI-KD) \cite{kim2022ai} integrates TF-KD and PS-KD and additionally employs a Generative Adversarial Network (GAN) to align distributions between sup-student and student.

Our work differs from the above work by designing a lightweight SKD framework with multi-source information fusion. We consider the more informative information from shallow layers of the networks and explore a consistent shape of model output for all classes of data.

\section{Preliminaries}
In this section, we present the preliminaries including the multi-class classification problem, KD, and SKD.
\subsection{Multi-class Classification}
Considering a supervised classification task on a training dataset $\mathcal{D}$, each data sample in the dataset is represented by $\left\{\bm{x},y\right\}\in\mathcal{D}$, where $\bm{x}$ indicates the input and $y$ is the corresponding label. We assume there are total $K$ classes such that $y\in\left\{1, \ldots, K\right\}$. We train a neural network model $h(\boldsymbol{\theta},\bm{x})$ parameterized by $\boldsymbol{\theta}$ to minimize the loss of a data sample on the model. 
A typical loss function for classification is cross-entropy loss. 
Denote $z:=h\left(\boldsymbol{\theta},\bm{x}\right)$ as the output logit of the model.
Applying the softmax function (with temperature $\tau=1$) to the model output, we can obtain the probability distribution $p$ for the input data $\boldsymbol{x}$:
\begin{align}
    p\left(z|\bm{x}\right) = \mathrm{softmax} \left(z,\tau\right) = \frac{\exp\left(z/\tau\right) }{\sum_{k=1}^{K}\exp(z_k/\tau)},
\end{align}
where $z_k$ indicate the $k$th element in $z$. When it is clear from the context, we use $p$ for short of $p\left(z|\bm{x}\right)$. The cross-entropy loss function is
\begin{equation}
\mathcal{L}_{\mathrm{CE}}(p\left(z|\bm{x}\right),y)=\frac{1}{K}\sum_{k=1}^{K}y_{k}\log p_{k},
\end{equation}
where $p_k$ indicates the $k$ th element of $p$.
The objective is to 
minimize the expected risk of the model on the whole dataset:
\begin{align}
    \min_{\boldsymbol{\theta}} \mathbb{E}_{\left\{\bm{x},y\right\}\in\mathcal{D}}\mathcal{L}_{\mathrm{CE}}(p\left(z|\bm{x}\right),y).
\end{align}

\subsection{Knowledge Distillation}
In KD, there exists another teacher model to guide the training of the target model, i.e., the student. A high temperature $\tau >1$ is applied to soften the model output probability distribution to facilitate transferring more knowledge from the teacher to the student \cite{hinton2015distilling}. 
Denote the output probability distribution  of the teacher with temperature $\tau$ for an input $\bm{x}$ by $q\left(z^{\prime}|\bm{x}\right)$, where $z^{\prime}$ is the output logit of the teacher.
The Kullback-Liebler (KL) divergence is employed to measure the difference between the teacher and student's model output probability distributions ($z^{\prime}$ and $z$):
\begin{equation}
\mathcal{L}_{\mathrm{KL}}\left(q\left(z^{\prime}|\bm{x}\right),p\left(z|\bm{x}\right)\right) = \frac{1}{K}\sum_{k=1}^{K} q_{k} \log \frac{q_{k}}{p_{k}}.
\end{equation}
Finally, the overall loss function for vanilla KD is:
\begin{equation}
\begin{split}
    \mathcal{L}_{\mathrm{KD}}\left(p,y,q\right) = &\mathcal{L}_{\mathrm{CE}}(p\left(z|\bm{x}\right),y) \\
    &+\tau^{2} \cdot \mathcal{L}_{\mathrm{KL}}\left(q\left(z^{\prime}|\bm{x}\right),p\left(z|\bm{x}\right)\right)
\end{split}
\end{equation}
The coefficient $\tau^{2}$ balances the cross-entropy and KL divergence losses when the temperature $\tau$ changes \cite{hinton2015distilling}.

\subsection{Self-Knowledge Distillation}
Self-knowledge distillation applies KD to improve model performance by utilizing the prior knowledge extracted from the model itself, which is different from the vanilla KD with a separate teacher model. To train the model $h\left(\boldsymbol{\theta},\boldsymbol{x}\right)$, we first extract some information $I\left(\boldsymbol{\theta},\boldsymbol{x}\right)$ from the model. 
$I\left(\boldsymbol{\theta},\boldsymbol{x}\right)$ may change with time, layers, and input data, but is not related to any other model. SKD executes a self-knowledge transfer (ST) loss to minimize the discrepancy between the model and the extracted information:
\begin{align}
\mathcal{L}_{\mathrm{ST}}\left(h\left(\boldsymbol{\theta},\boldsymbol{x}\right),I\left(\boldsymbol{\theta},\boldsymbol{w}\right)\right) := \rho\left(h\left(\boldsymbol{\theta},\boldsymbol{w}\right),I\left(\boldsymbol{\theta},\boldsymbol{x}\right)\right), 
\end{align}
where $\rho$ is a metric function, which varies for different SKD methods. 
For example, $\rho$ corresponds to a $l$2-norm in BYOT, the KL Divergence in PS-KD, and the adversarial loss in AI-KD, etc. The ST loss function may take effect at different parts of the model $h\left(\boldsymbol{\theta},\boldsymbol{w}\right)$. For example, the ST loss function updates the shallow layers of the model in BYOT and updates the whole model in TW-SKD and IC-SKD.
Overall, the SKD loss function combines the original loss function using the hard labels and the ST loss function:
\begin{equation}
    \mathcal{L}_{\mathrm{SKD}} =\mathcal{L}_{\mathrm{CE}}(p\left(z|\bm{x}\right),y) +\zeta \cdot \mathcal{L}_{\mathrm{ST}}\left(h\left(\boldsymbol{\theta},\boldsymbol{x}\right),I\left(\boldsymbol{\theta},\boldsymbol{x}\right)\right)
\end{equation}
where $\zeta$ measures the importance of ST loss, which may vary for different SKD methods.

\section{Proposed Methods}
In this section, we propose our DRG and DSR methods to achieve multi-source information fusion for SKD performance improvement.

\subsection{Distillation with Reverse Guidance (DRG)}
\textbf{Motivation:} Different layers in a neural network grab different features of the input data. Typically, shallower layers can capture more
shape and edge information while deeper layers can learn more
detailed semantic information. 
The shape and edge feature of the input data vanishes gradually as the layers become deepen, resulting in ignorance of edge information in the final model output and severe model overfitting. 
 By adding an AC to a shallow layer, we can construct a teacher model for the original model. The output of the AC is usually more underfitting than the whole model as it has a smaller model architecture. Related works have revealed the effectiveness of a ``poor'' teacher for KD \cite{yuan2020revisiting}. However, they have neglected the potential of shallow layers for guiding the training of the whole model. 
Thus, we propose to use the shallow layer to \textit{reversely} guide the training of the whole model to achieve information fusion of both edge and detailed features of date.

\textbf{DRG design: } The framework of DRG is demonstrated on the left-hand side of Fig.~\ref{fig:methods}. We consider neural networks with sequential layers/blocks, such as ResNet \cite{he2016deep}.
DRG introduces an add-on structure, i.e., AC, to the output of a shallow layer/block \footnote{Note that we just additionally train one AC in DRG, as compared to one AC for each block in BYOT. In this sense, our DRG is a lightweight framework. We will discuss the selection of the position of shallow layers in Section V and show that one AC is enough for DRG to achieve a remarkable SKD performance.}, constructing a ``poor'' teacher.  Let $\boldsymbol{w}$ be the parameter of the AC. The  teacher model can be represented by $g\left(\hat{\boldsymbol{\theta}},\boldsymbol{w},\boldsymbol{x}\right)$, where $\hat{\boldsymbol{\theta}}\subset \boldsymbol{\theta}$ is the parameter of the easier layers of the whole model before the layer connected to the AC. Denote the output logit and corresponding output probability distribution of $g\left(\hat{\boldsymbol{\theta}},\boldsymbol{w},\boldsymbol{x}\right)$ taking $\bm{x}$ as input by $z':=g\left(\hat{\boldsymbol{\theta}},\boldsymbol{w},\boldsymbol{x}\right)$ and $q\left(z^{\prime}|\bm{x}\right):=\mathrm{softmax}\left(z',\tau\right)$, respectively. We use the cross-entropy loss function to train the ``poor'' teacher model and the whole model simultaneously using the following hard-label loss. 
\begin{equation}
\label{HL}
\begin{split}
\mathcal{L}_{\mathrm{HL}}= \mathcal{L}_{CE}(q(z^{\prime}|\bm{x}),y) + \mathcal{L}_{CE}(p(z|\bm{x}),y).
\end{split}
\end{equation}
To achieve reverse guidance, the ``poor'' teacher guides the whole model training by minimizing the KL divergence:
\begin{equation}
\label{RI}
\begin{split}
\mathcal{L}_{\mathrm{RG}}=\tau^{2} \cdot \mathcal{L}_{KL}(q(z^{\prime}|\bm{x}),p(z|\bm{x}))
\end{split}
\end{equation}
Overall, the whole loss function of DGR is  
\begin{equation}
\label{DRG}
    \mathcal{L}_{\mathrm{DRG}} =\mathcal{L}_{\mathrm{HL}} +\alpha \cdot  \mathcal{L}_{\mathrm{RG}},
\end{equation}
where $\alpha$ is a coefficient between two losses.

Algorithm \ref{alg:alg1} demonstrates the model training process of DRG, where $\gamma$ denotes the learning rate and $T$ indicates the total number of training iterations. $\boldsymbol{\theta}^t$ and $\boldsymbol{w}^t$ represent the model parameters at iteration $t$. In each iteration $t$, a mini-batch of data $B^t\subset\mathcal{D}$ is randomly sampled to train the model. The mini-batch is simultaneously fed into the model (line \ref{algline:feed-model}) and the teacher, which is constructed by shallow layers of the model and the AC,  (line \ref{algling:feed-teacher}). Based on the output of the original model and the teacher, we calculate the DRG loss, i.e., $\mathcal{L}_{\mathrm{DRG}}$ (line \ref{algline:loss}) and update the model and auxiliary parameters (line \ref{algline:sgd-o} - \ref{algline:sgd-w}) according to SGD.

\begin{algorithm}[t]
\caption{Distillation with Reverse Guidance (DRG).}\label{alg:alg1}
\begin{algorithmic}[1]
\renewcommand{\algorithmicrequire}{{\textbf{Input: }} $\mathcal{D}, \gamma, \tau, \alpha, T$}
\REQUIRE
\STATE Initialize $\boldsymbol{\theta}\leftarrow\boldsymbol{\theta}^0$, $\boldsymbol{w}\leftarrow \boldsymbol{w}^0$;
\FOR {$t \in \left\{0,\ldots,T-1\right\}$}
\STATE Randomly sample $B^t$  from $D$;
\STATE $z\leftarrow h(\boldsymbol{\theta}^t,B^t)$; \label{algline:feed-model}
\STATE Compute loss $L_{\mathrm{HL}}$ using (\ref{HL}); 
\STATE $z'\leftarrow g\left(\hat{\boldsymbol{\theta}}^t,\boldsymbol{w}^t,B^t\right)$;\label{algling:feed-teacher}
\STATE Compute discrepancy $L_{\mathrm{RG}}$ using (\ref{RI});
\STATE Compute loss $L_{\mathrm{DRG}}$ using (\ref{DRG}); \label{algline:loss}
\STATE $\boldsymbol{\theta}^{t+1} \xleftarrow{} \theta^{t}- \gamma \cdot \nabla L_{\mathrm{DRG}}$; \label{algline:sgd-o}
\STATE $\boldsymbol{w}^{t+1} \xleftarrow{} w^{t}- \gamma \cdot \nabla L_{\mathrm{DRG}}$; \label{algline:sgd-w}
\ENDFOR
\end{algorithmic}
\label{alg1}
\end{algorithm}
\subsection{Distillation with Shape-wise Regularization (DSR)}
\begin{figure}[t!]
\begin{center}
\centerline{\includegraphics[width=1\columnwidth]{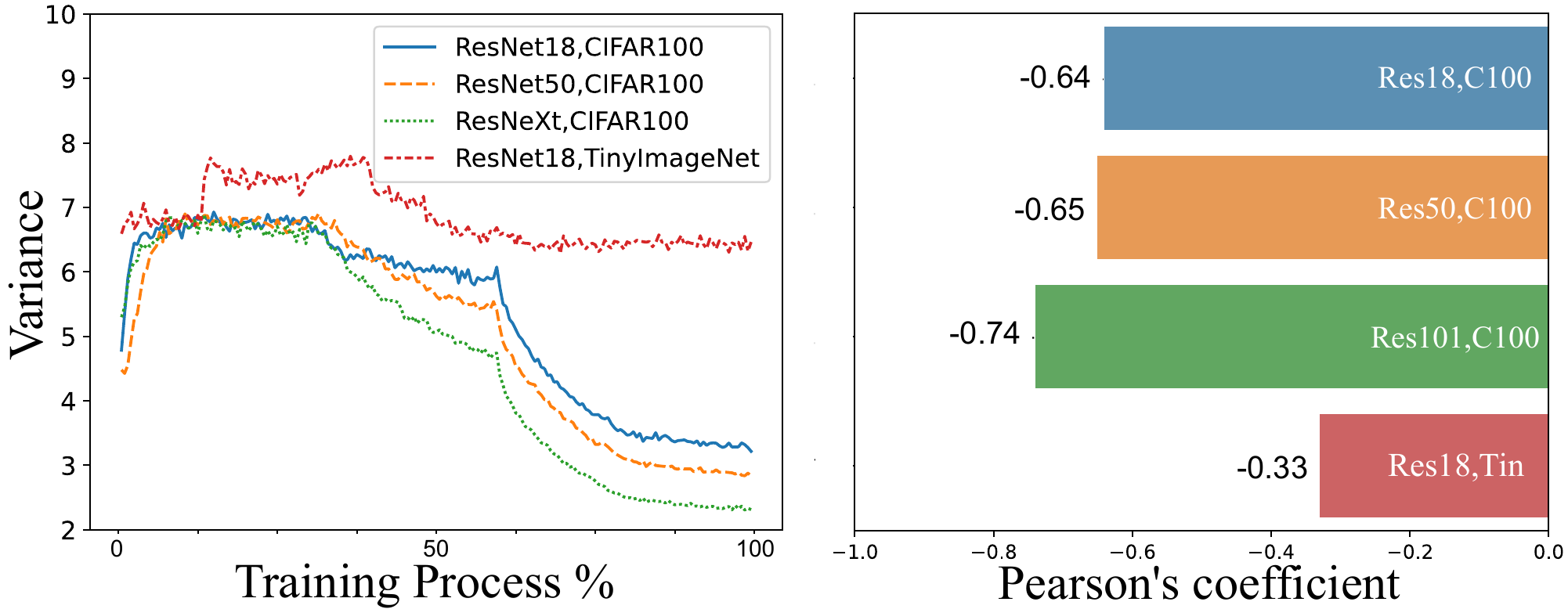}}
\caption{The variance of ranked outputs in one epoch along the training process (left) and Pearson's coefficient of variance and accuracy (right) for different datasets trained with various models. ResNet, CIFAR100, TinyImageNet are abbreviated as ``Res'', ``C100'', and ``Tin'' respectively. }
\label{fig:pearson}
\end{center}
\end{figure}

\textbf{Motivation:} 
Existing works have investigated the intra-class similarity of input data, such as CS-KD and DDGSD. 
However, to the best of our knowledge, no work has stressed the consistent property of model output among different classes, i.e., inter-class similarity. To illustrate the necessity of exploring a consistent model output property of different classes of data, we evaluate the variance of ranked model outputs, as demonstrated on the left-hand side of Fig. \ref{fig:pearson}. Ranking outputs\cite{liang2022efficient} according to class probability would eliminate the class inharmony and gives more concentration on the overall interaction between classes.  We train CIFAR100 and TinyImageNet datasets using various models till converge and normalize the training time to the training process. The variance is calculated by taking an average of the variances of each element in the model outputs over all test data samples. We can observe that the variance of ranked model output decreases along with the training process, which corresponds to increasing model accuracy. On the right-hand side of Fig. \ref{fig:pearson}, we calculate the Pearson coefficients between model accuracy and the variance of ranked model output for different datasets trained with various models. All results exert a strong negative relation between model accuracy and ranked output variance. This implies that along with model training, the model outputs from various classes have a consistent tendency after being ranked. The phenomenon motivates us to regularize the ranked model output shape of different input data to improve the performance of SKD.

\textbf{DSR design: }The framework of DRG is demonstrated on the right-hand side of Fig.~\ref{fig:methods}. In each iteration $t$, we rank the elements in the model output according to the non-decreasing order and obtain $\tilde{z}^t=\{\tilde{z}^{t}_{1}, \tilde{z}^{t}_{1},\cdots\tilde{z}^{t}_{K}\}$, such that $\tilde{z}^{t}_{1}\leq\tilde{z}^{t}_{1}\leq\ldots\leq\tilde{z}^{t}_{K}$. DSR achieves the consistency of ranked model output between different input date leveraging the ranked model output of the last iteration, i.e., $\tilde{z}^{t-1}$. We use KL divergence to regularize the model using  $\tilde{z}^{t-1}$, defined as the $\mathcal{L}^t_{\mathrm{SR}}$:
\begin{equation}
\label{SR}
    \mathcal{L}_{\mathrm{SR}}^{t} =\tau^{2} \cdot \mathcal{L}_{\mathrm{KL}}(p(\tilde{z}^{t-1}|\bm{x}),p(\tilde{z}^{t}|\bm{x})).
\end{equation}
Overall, DSR combines the vanilla classification loss and $\mathcal{L}^t_{\mathrm{SR}}$ for SKD model training: 

\begin{equation}
\label{DSR}
    \mathcal{L}_{\mathrm{DSR}} =\mathcal{L}_{\mathrm{CE}}(p(z|\bm{x}),y) +\beta \cdot \mathcal{L}_{\mathrm{SR}}^t,
\end{equation}
where $\beta$ measures the importance of $\mathcal{L}^t_{\mathrm{SR}}$ compared to the original classification loss.

Algorithm \ref{alg:alg2} shows the training process of DSR. Specifically in each iteration $t$, data batch $B^{t}$ is randomly sampled to train the model. Outputs of the model, i.e., $z$, are then ranked in ascending order  to obtain $\tilde{z}$(line 5). The DSR loss is computed using the ranked model output in the last iteration, i.e., $\tilde{z}^{t-1}$, (line 6). After updating the model parameters with SGD (line 7), $\tilde{z}^{t}$ will be recorded (line 8) and used in the next iteration.

\begin{algorithm}[t]
\caption{Distillation with Shape-wise Regularization.}\label{alg:alg2}
\begin{algorithmic}[1]
\renewcommand{\algorithmicrequire}{{\textbf{Input: }}untrained single model $S(\theta)$, training data $D$}
\renewcommand{\algorithmicrequire}{{\textbf{Input: }} $\mathcal{D}, \gamma, \tau, \beta, T$}
\REQUIRE
\STATE Initialize $\boldsymbol{\theta}\leftarrow\boldsymbol{\theta}^0, \tilde{z}^{-1}\leftarrow \boldsymbol{0}$;
\FOR {$t \in \left\{0,\ldots,T-1\right\}$}
\STATE Randomly sample $B^t$  from $D$;
\STATE $z\leftarrow h(\boldsymbol{\theta}^t,B^t)$; \label{algline:feed-model-2}
\STATE Rank $z$ in ascending order to obtain $\tilde{z}$;\label{algline:rank}
\STATE Compute loss $L_{\mathrm{DSR}}^t$ using (\ref{DSR}); 
\STATE $\boldsymbol{\theta}^{t+1} \xleftarrow{} \theta^{t}- \gamma \cdot \nabla L_{\mathrm{DSR}}$; 
\STATE Store $\tilde{z}$ for the next iteration; 
\label{algline:sgd-2}
\ENDFOR
\end{algorithmic}
\label{alg2}
\end{algorithm}

We can combine our DRG and DSR methods for SKD using the following overall loss function:
\begin{equation}
    \mathcal{L}=\mathcal{L}_{\mathrm{HL}} +\alpha \cdot  \mathcal{L}_{\mathrm{RG}} +\beta \cdot \mathcal{L}_{\mathrm{SR}}^t.
\end{equation}

\begin{table*}[ht!]
\caption{Top-1 test accuracy on CIFAR100. Values marked in \textcolor[rgb]{0.95,0.1,0.1}{Red}, \textcolor{blue}{Blue} are the best and the second best accuracy respectively. }\label{table:cifar}
\begin{center}
\begin{small}
\begin{sc}
\renewcommand\arraystretch{1.1}
\setlength{\tabcolsep}{1.5mm}{
\begin{tabular}{lccccccr}
\hline
Methods & ResNet18 & ResNet50 & ResNet101 & ResNeXt50\_32x4d & DenseNet-121\\
\hline
Vanilla    & 77.29\% & 77.07\%&  78.52\%&      78.87\%&      78.70\%&\\
BYOT    & 78.25\%& 79.63\%& 80.71\%&   80.18\%&  79.63\%&\\
CS-KD    & 78.55\%& 76.91\%&   77.43\%&      79.69\%&      78.92\%&\\
PS-KD    & 78.67\%& 79.02\%&   79.41\%&      80.38\%&      79.52\%&\\
DLB    & \textcolor[rgb]{0.95,0.1,0.1} {79.52\%}& \textcolor{blue} {79.88\%}&  80.02\% &      80.52\%&             79.64\%&\\
ZF-LS$\rm{_{lb}}$     & 77.49\%& 77.38\%&   77.27\%&      79.42\%&      78.87\%&\\
TF-KD$\rm{_{reg}}$      & 78.33\%& 78.30\%&   79.19\%&      79.27\%&      79.38\%&\\
\hline
DRG (ours)     & 79.07\% (+1.78\%)& 79.87\% (+2.80\%)&  \textcolor[rgb]{0.95,0.1,0.1} {80.86\% (+2.34\%)} &  \textcolor[rgb]{0.95,0.1,0.1} {81.01\% (+2.14\%)}&   \textcolor[rgb]{0.95,0.1,0.1} {79.99\% (+1.29\%)}   &\\
DSR (ours)   & 78.15\% (+0.88\%)& 79.12\% (+2.05\%)&   79.78\% (+1.26\%)&   79.01\% (+0.14\%)&  79.08\% (+0.38\%)   &\\
DRG+DSR (ours)   & \textcolor{blue} {79.30\% (+2.01\%)}& \textcolor[rgb]{0.95,0.1,0.1} {79.94\% (+2.87\%)}&  \textcolor{blue} {80.72\% (+2.20\%)} &   \textcolor{blue} { 80.91\% (+2.04\%)}   &  \textcolor{blue} {79.76\%  (+1.26\%)}&\\
\hline
\end{tabular}}
\end{sc}
\end{small}
\end{center}
\end{table*}
\begin{table*}[ht!]
\caption{Top-1 test accuracy on TinyImageNet. Values marked in \textcolor[rgb]{0.95,0.1,0.1}{Red}, \textcolor{blue}{Blue}  are the best and the second best accuracy respectively. }\label{table:tinyimagenet}
\begin{center}
\begin{small}
\begin{sc}
\renewcommand\arraystretch{1.1}
\setlength{\tabcolsep}{2.5mm}{
\begin{tabular}{lccccr}
\hline
Methods & ResNet18 & ResNet50 & ResNeXt50\_32x4d\\
\hline
Vanilla    & 56.69\% & 58.07\%&  59.55\%\\
BYOT    & \textcolor{blue} {57.69\%}& 60.59\%&   60.07\%\\
PS-KD    & 57.05\%& \textcolor{blue} {60.70\%}&   60.87\% \\
DLB    & 57.09\%& 59.89\%& 60.65\%\\
\hline
DRG (ours)      & 57.57\% (+0.88\%)& 60.41\% (+2.34\%)&  \textcolor{blue} {60.94\% (+1.39\%)} \\
DSR (ours)   & 56.75\% (+0.06\%)& 58.34\% (+0.27\%)& 60.34\% (+0.79\%)\\
DRG+DSR (ours)   & \textcolor[rgb]{0.95,0.1,0.1} {58.08\% (+1.39\%)}& \textcolor[rgb]{0.95,0.1,0.1} {61.04\% (+2.97\%)}&  \textcolor[rgb]{0.95,0.1,0.1} {61.14\% (+1.59\%)} \\
\hline
\end{tabular}}
\end{sc}
\end{small}
\end{center}
\end{table*}

\section{Experiments}
We conduct experiments for our proposed method over various datasets and models. 
First, we introduce settings including datasets, models, baselines, etc. Then, we analyze the experimental results for different datasets. Our code is available at \url{https://github.com/xucong-parsifal/LightSKD}.

\subsection{Settings}
\textbf{Datasets. }We employ five datasets for classification tasks, i.e., CIFAR100, TinyImageNet, Caltech101, Stanford Dogs and  CUB200.
\begin{itemize}
    \item \textbf{CIFAR100: }CIFAR100\cite{krizhevsky2009learning} is a classical 100-class classification dataset. It contains 50,000 images for training and 10,000 for test. The image size is 32x32 pixels.
    \item \textbf{TinyImageNet:} TinyImageNet is a subset of ImageNet\cite{deng2009imagenet}, with 100, 000 train data samples and 10,000 test samples. There are 200 classes in total. The size of an image is 32x32 pixels.
    \item \textbf{Caltech101}: Caltech101 is a large tough-grained dataset for classification and object detection. There are 101 main classes and 1 background class in total. 
    \item \textbf{Stanford Dogs / CUB200:} Stanford Dogs and CUB200 are large fine-grained datasets that consist of 120 dog classes and 200 bird classes, respectively. 
\end{itemize}
In all experiments, training samples are processed with RandomCrop (32x32 for CIFAR100,TinyImageNet; 224x224 for others) and RandomHorizontalFlip to ensure  that all images have a consistent size and to add randomness to the training process. 

\textbf{Models.} We employ five classical neural network models for the above datasets including ResNet18, ResNet50, ResNet101 \cite{he2016deep}, ResNeXt50\_32x4d \cite{xie2017aggregated}, and DenseNet121\cite{huang2017densely}. The ResNet series is well-known for its innovative shortcut connections, which help to reduce overfitting. In contrast, the DenseNet architecture was the first to introduce fully-connected blocks as a means of improving feature reuse and facilitating information flow between layers.

\textbf{Environment and hardwares}: Our implementations are based on PyTorch, with Python version 3.8.5, Torch version 1.13.0, and Torchvision version 0.14.0. All experiments were conducted using an NVIDIA RTX 3090 with 24GB memory. 

\textbf{Hyperparameters.} 
We fixed the number of epochs to 200 and set the temperature $\tau$ using a grid search. We set hyperparameters $\alpha$ and $\beta$ to 0.2 and 1, respectively, and employ a manual learning rate adjustment mechanism for our experiments. For CIFAR100, the initial learning rate was set to 0.1 and decreased to 0.2 of its previous value at 60, 120, and 160 epochs. For TinyImageNet, Stanford Dogs, CUB200, and Caltech101, the initial learning rate was set to 0.1 and decreased to 0.1 of its previous value at 100 and 150 epochs. We use a batch size of 128 for CIFAR100 and TinyImageNet, and 64 for the other datasets. The optimizer used was SGD with a momentum of 0.9 and weight decay of 5e-4. 
For DRG, we add an AC after the second block of the model to construct the ``poor'' teacher.

\textbf{Baselines.} We combine our proposed method with the following methods:
\begin{itemize}
    \item Vanilla: training the original model without SKD;
    \item BYOT \cite{zhang2019your}: adding an Auxiliary Classifier (AC) to each block of the model;
    \item CS-KD \cite{yun2020regularizing}: an IC-SKD method that uses two batched of data samples from the same class and minimizes the output discrepancy between the two batches;
    \item PS-KD \cite{kim2021self}: a TW-SKD method that employs the model in the last epoch as a teacher;
    \item DLB\cite{shen2022self}: a TW-SKD method that regards the model in the last iteration as a teacher, meanwhile employing different augmentation techniques for the same data batch. It differs from PS-KD in the supervision granularity and data preprocessing. 
    \item ZF-LS${\rm _{lb}}$\cite{liang2022efficient}: a label smoothing method that minimizes the cross entropy between the ranked model outputs and zipf's distribution;
    \item TF-KD${\rm _{reg}}$ \cite{yuan2020revisiting}: an SKD based on ameliorating LSR.
\end{itemize}

\begin{figure}[h]
\begin{center}
\centerline{\includegraphics[width=1\columnwidth]{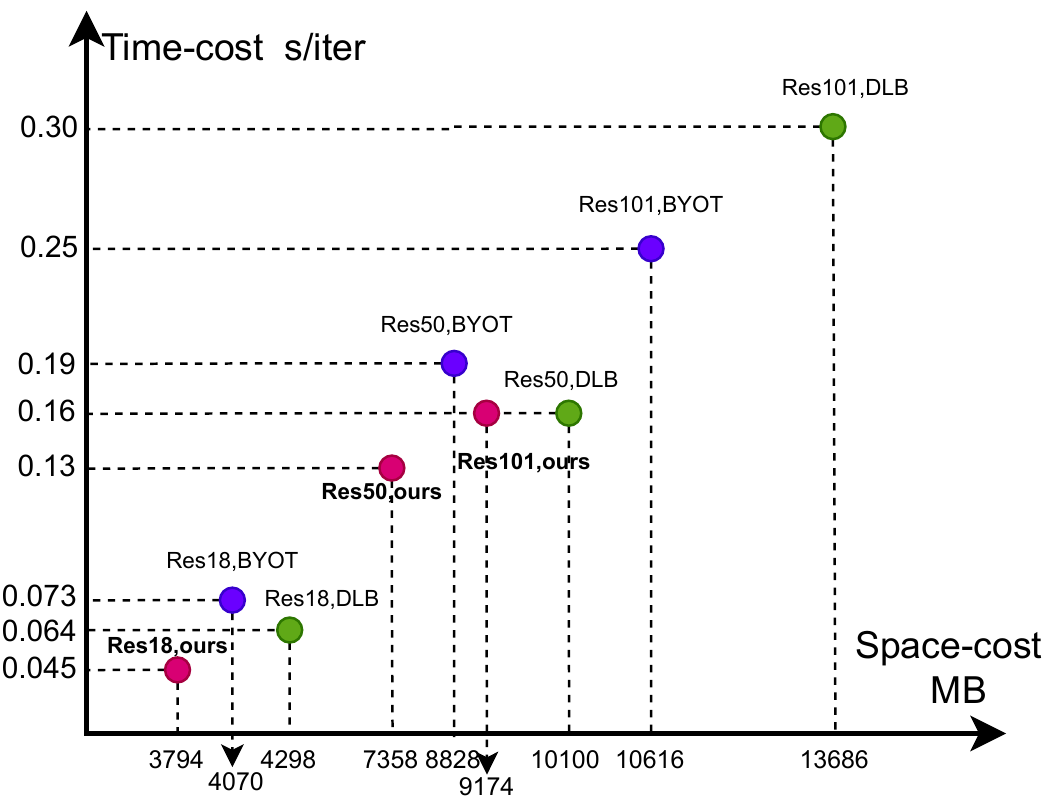}}
\caption{Time and space cost of different methods trained with various models on  CIFAR100. ResNet is abbreviated as ``Res''. \textcolor{blue}{Blue}, \textcolor{green}{Green} and \textcolor{red}{Red} points represent experiments of BYOT, DLB and our methods respectively.}
\label{fig:times}
\end{center}
\end{figure}

\begin{figure*}[h]
\begin{center}
\centerline{\includegraphics[width=1.83\columnwidth]{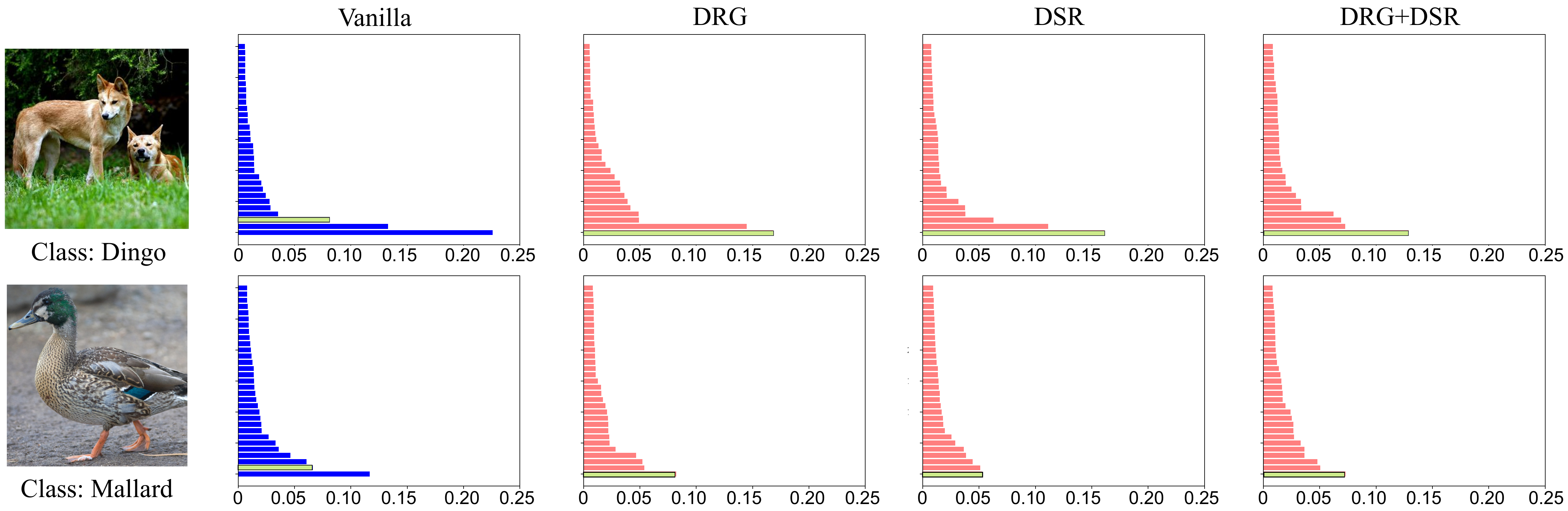}}
\caption{Example experimental results on Stanford Dogs (top) and CUB200 (bottom). All bar figures show the ranked predictive probability of the top 30 classes, with ground-truth (GT) classes marked in \textcolor[rgb]{0,0.8,0}{Green}. The baseline results for vanilla single model training are shown in the second column, while the other columns display results from DRG and DSR, and their combination. In the face of complex tasks, our results show lower probabilities for the GT class and higher probabilities for other classes. This suggests that our methods extract more integrated information and are less overconfident and overfitting, resulting in a more careful and delicate decision-making process.
}
\label{fig:example}
\end{center}
\end{figure*}

\subsection{Experimental results}

\subsubsection{Results on CIFAR100 and TinyImageNet}
Our results are presented in Table \ref{table:cifar} (for CIFAR100) and Table \ref{table:tinyimagenet} (for TinyImageNet).

Compared with baseline algorithms, we have the following observations:
\begin{itemize}
\item \textbf{Compared with vanilla single model training:} our methods consistently outperform the  vanilla single model training in  top-1 accuracy, with a significant improvement ranging from 1.26\% to 2.87\%.

\item \textbf{Compared with BYOT, CS-KD, PS-KD, ZF-LS, and TF-KD}: Our methods generally achieve higher accuracy than these methods, with an average accuracy boost of 1.08\%. Particularly for CIFAR100 over ResNet18 model, our methods exceed their maximum accuracy by 0.97\%.

\item  \textbf{Compared with DLB}: To the best of our knowledge, DLB is the current claimed SOTA. Our results show that our methods perform better than DLB. Especially, our methods surpass DLB on large-scale networks, such as the ResNet100 for CIFAR100. This is because DLB uses the same images with different transformations, which may lead to overfitting and diluting the regularization effects in larger networks. Our methods avoid this problem.
\end{itemize}

Notably, the combination of our methods, i.e., DRG+DSR, is particularly effective and has achieved SOTA performance. Although DSR may not individually achieve SOTA, it has contributed significantly to the success of the combination (+0.51\% on ResNet18, TinyImageNet; +0.63\% on ResNet18 and TinyImageNet), surpassing its individual accuracy boost. 

\textbf{Time and space costs.} The time and space costs of different methods on CIFAR100 dataset with various models are shown in Fig. \ref{fig:times}, where the time cost is evaluated by the consuming time of each iteration and the space cost is the storage space of the models. We can observe that BYOT takes about 0.064s per iteration on ResNet18 and spends much more when the model gets larger. Although DLB is faster than BYOT on small models, it incurs a vast time cost on ResNet101, which may result from re-sampling the training dataset to construct mini-batches and frequently recording the images and outputs of the last iteration. Remarkably, our combined method DRG+DSR receives the least time and space cost. Specifically, the time cost of our DRG+DSR is about only 70 percent of that of others; the Space-cost of our DRG+DSR is also extraordinarily smaller than others ($\times$0.67 $\sim$ $\times$0.83). Most importantly, we can achieve better performance than BYOT and DLB.

\textbf{Robustness.}
Our proposed methods are more robust over different neural network models than baselines. Specifically for CIFAR100, we achieve the best results among all methods, especially for large-scale models such as ResNet100, ResNeXt50\_32$\times$4d, and DenseNet-121, indicating the robustness of our methods across different models. 

\subsubsection{Results on large-scale fine-grained datasets}
We extend our experiments to include the large fine-grained datasets of Stanford Dogs and CUB200. Figure \ref{fig:example} shows the ranked model output probability of the top 30 classes for two data examples.
The Green bars mark the ground-truth label. 
Our results indicate that vanilla training of a single model may give a wrong prediction as the predicted label with the highest probability is not consistent with the true label.  
In comparison, our methods generate model output probability with low variance, exerting  higher probabilities for several classes outside the true label. This means our models could select a range of candidate classes and make decisions more carefully and delicately, rather than making an exact decision that neglects the relationships between different classes.

\subsubsection{Compatibility Analysis}
To validate the effectiveness and compatibility of our methods over the existing methods, we plug DSR and DRG into Cutout \cite{devries2017improved} and  PS-KD. 
Cutout is a popular data augmentation technique, which employs a mask to randomly eliminate part of the input image data. We set the number of masks and mask size to 1 and 16px respectively, which is consistent with \cite{shen2022self}.

Table \ref{table:compatibility} presents the performance of these methods before and after the integration of DRG, DSR, and their combination.
The results demonstrate that the addition of DRG, DSR, or their combination significantly improves the accuracy of Cutout by 0.82\% to 2.42\%. Similarly, the integration of these methods with PS-KD results in an accuracy boost of 0.39\% to 0.71\% compared to vanilla PS-KD.

\begin{table}[t]
\caption{Results of different combinations of our methods and existing methods for CIFAR100 over ResNet18. }\label{table:compatibility}
\begin{center}
\begin{small}
\begin{sc}
\renewcommand\arraystretch{1.1}
\setlength{\tabcolsep}{4mm}{
\begin{tabular}{lcr}
\hline
Methods & Accuracy \\
\hline
Cutout & 77.39\%\\  
Cutout+DRG & 80.12\%(+2.73\%)\\
Cutout+DSR & 78.21\%(+0.82\%)\\
Cutout+DRG+DSR & 79.81\%(+2.42\%)\\
PS-KD & 78.67\%\\
PS-KD+DRG & 79.18\%(+0.51\%)\\
PS-KD+DSR & 79.38\%(+0.71\%)\\
PS-KD+DRG+DSR & 79.06\%(+0.39\%)\\
\hline
\end{tabular}}
\end{sc}
\end{small}
\end{center}
\end{table}

\begin{table}[t]
\caption{Accuracy and time-cost of different block subsets in DRG for CIFAR100 over ResNet18.}\label{table:DRG_ablation}
\begin{center}
\begin{small}
\begin{sc}
\renewcommand\arraystretch{1.1}
\setlength{\tabcolsep}{1mm}{
\begin{tabular}{lcccccr}
\hline
B \#1 & B \#2 & B \#3 & Accuracy \% & Time-Cost (s/iter).\\
\hline
\color[rgb]{0,0.8,0}{\ding{52}}   &  &  &  78.93\%($\times0.9953$)& 0.044($\times0.99$)  \\
 &\color[rgb]{0,0.8,0}{\ding{52}}  &  &  79.30\%&  0.045 \\
   &  & \color[rgb]{0,0.8,0}{\ding{52}} &  76.96\%($\times0.9705$)& 0.046(x1.01)  \\
\color[rgb]{0,0.8,0}{\ding{52}}  & \color[rgb]{0,0.8,0}{\ding{52}} &  &  79.42\%($\times 1.0015$)&  0.052($\times 1.17$) \\
\color[rgb]{0,0.8,0}{\ding{52}}   &  & \color[rgb]{0,0.8,0}{\ding{52}} &  78.54\%($\times 0.990$)& 0.054($\times 1.21$)    \\
  & \color[rgb]{0,0.8,0}{\ding{52}} & \color[rgb]{0,0.8,0}{\ding{52}}  &  79.32\%($\times 1.0002$)&  0.055($\times 1.22$) \\
\hline
\end{tabular}}
\end{sc}
\end{small}
\end{center}
\end{table}

\begin{figure}[t]
\begin{center}
\centerline{\includegraphics[width=0.80\columnwidth]{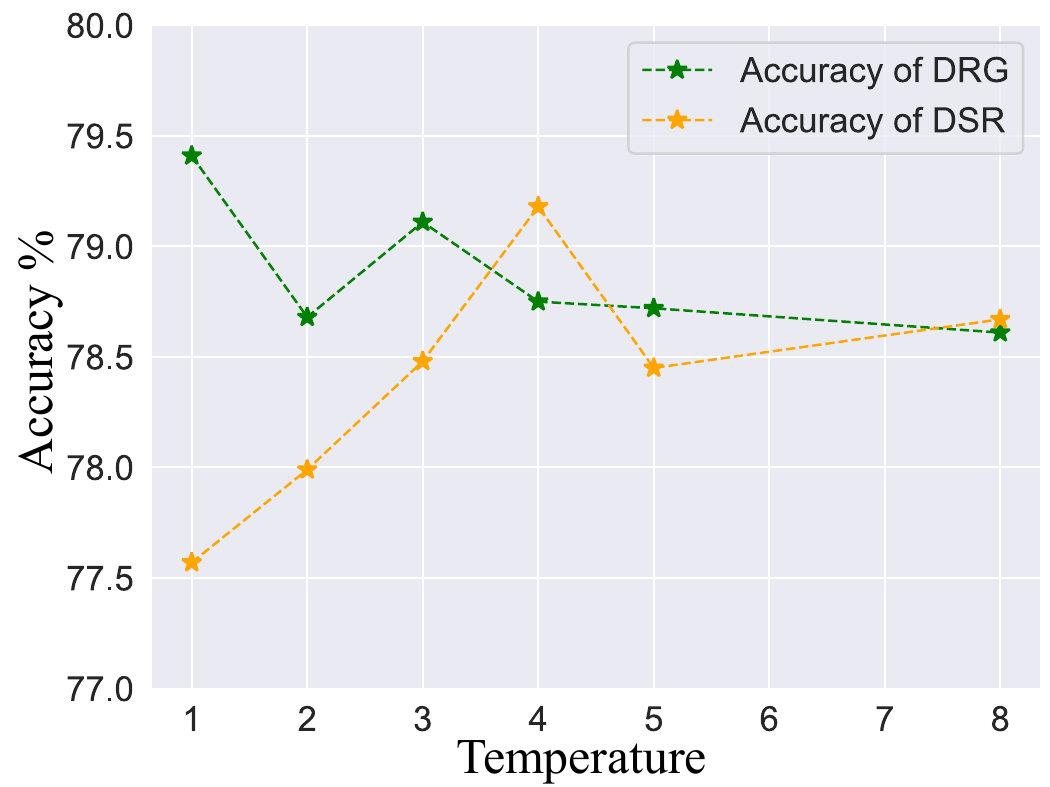}}
\caption{Performance of DRG and DSR under varying temperature. }
\label{fig:temperature}
\end{center}
\end{figure}

\begin{figure}[t]
\begin{center}
\centerline{\includegraphics[width=1\columnwidth]{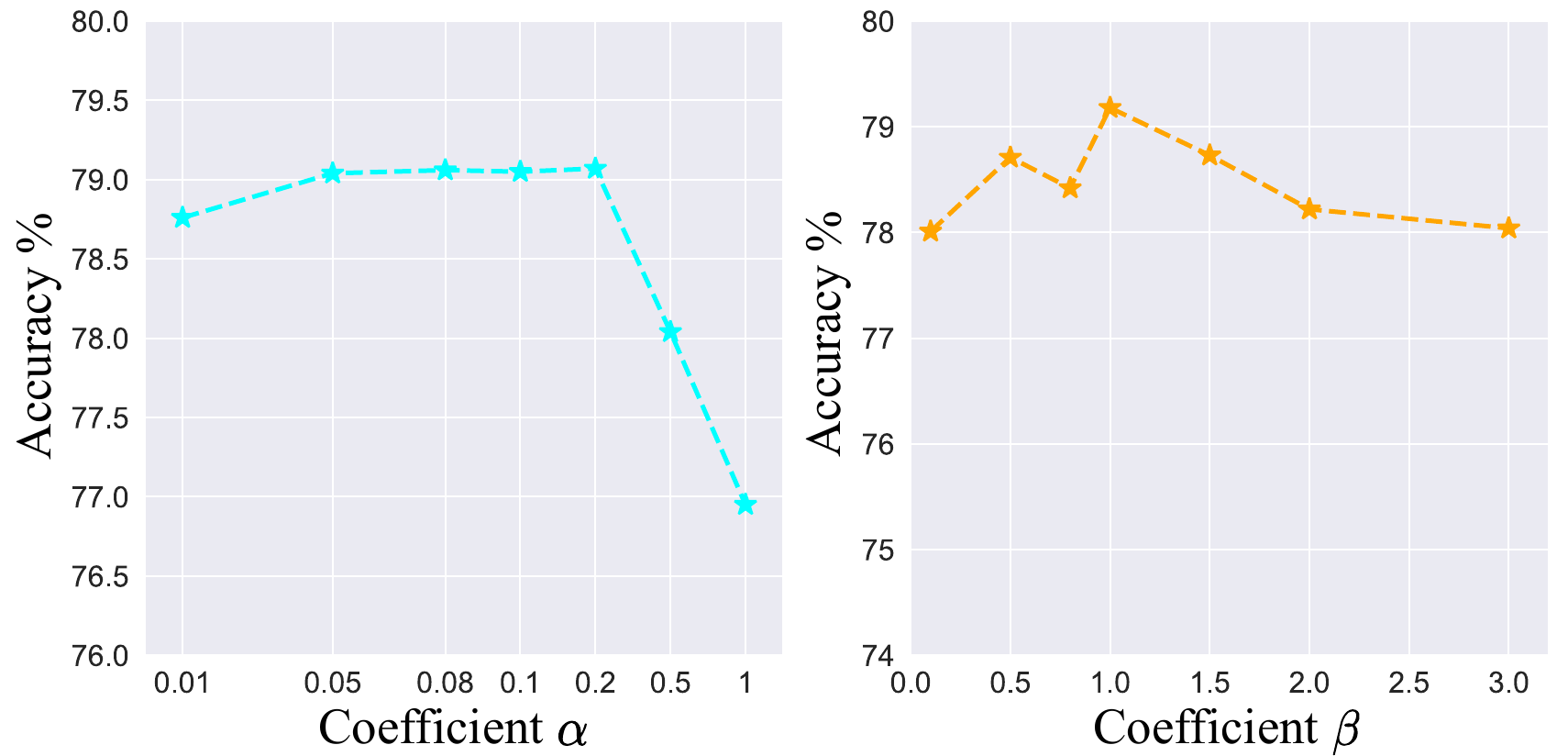}}
\caption{Accuracy under varying $\alpha$ and $\beta$. }
\label{fig:coefficients}
\end{center}
\end{figure}

\section{Ablation Study}

In this section, we conduct an ablation study of our proposed methods. We first explore the number of teachers and the position of selected blocks in DRG. Then we evaluate the effect of different hyperparameters including temperature and the coefficients in objective loss functions.

\subsection{AC number and block position in DRG}
For DRG, we can choose one or a subset of blocks in the neural network model to add ACs to selected blocks in order to accelerate the model learning process while maintaining accuracy. Table \ref{table:DRG_ablation} displays the accuracy and time cost results of CIFAR100, over the ResNet18 model for different sets of selected blocks.

We have the following observation:
\begin{itemize}
    \item  When adding AC to a single block in the deeper layer of the model, such as the third block (B $\#$3) compared to the first and second blocks, DRG experiences a sharp decrease in test accuracy. That is because the outputs of deeper layers have a higher level of similarity with the final output, contributing less to the formulation fusion and possibly leading to overfitting. 
    \item Selecting two blocks to add ACs show subtle accuracy improvement with a significantly increased time cost. Therefore, only constructing one ``poor'' teacher is enough for our DRG, resulting in a lightweight SKD design.
\end{itemize}

\subsection{Hyperparameters}

\textbf{Temperature $\tau$:} We evaluate the performance of DRG and DSR under varying temperature on CIFAR100, ResNet18, as shown in Fig. \ref{fig:temperature}. The results indicate that DRG and DSR achieve the highest accuracy when the temperature are set to 1 and 4, respectively. 

\textbf{Coefficients $\alpha$ and $\beta$:} We evaluate the performance of DRG and DSR for different coefficients $\alpha$ and $\beta$ in \eqref{DRG} and \eqref{DSR} on CIFAR100, ResNet18. We vary $\alpha$ and $\beta$ from 0.01 to 1 and  from 0.1 to 3, respectively. The results in Fig. \ref{fig:coefficients} show that the best accuracy is achieved when $\alpha$ and $\beta$ are set to 0.2 and 1, respectively.
This suggests that a moderate level of usage of both DRG and DSR provides optimal performance for SKD.

\begin{figure*}[t]
\begin{center}
\centerline{\includegraphics[width=1.6\columnwidth]{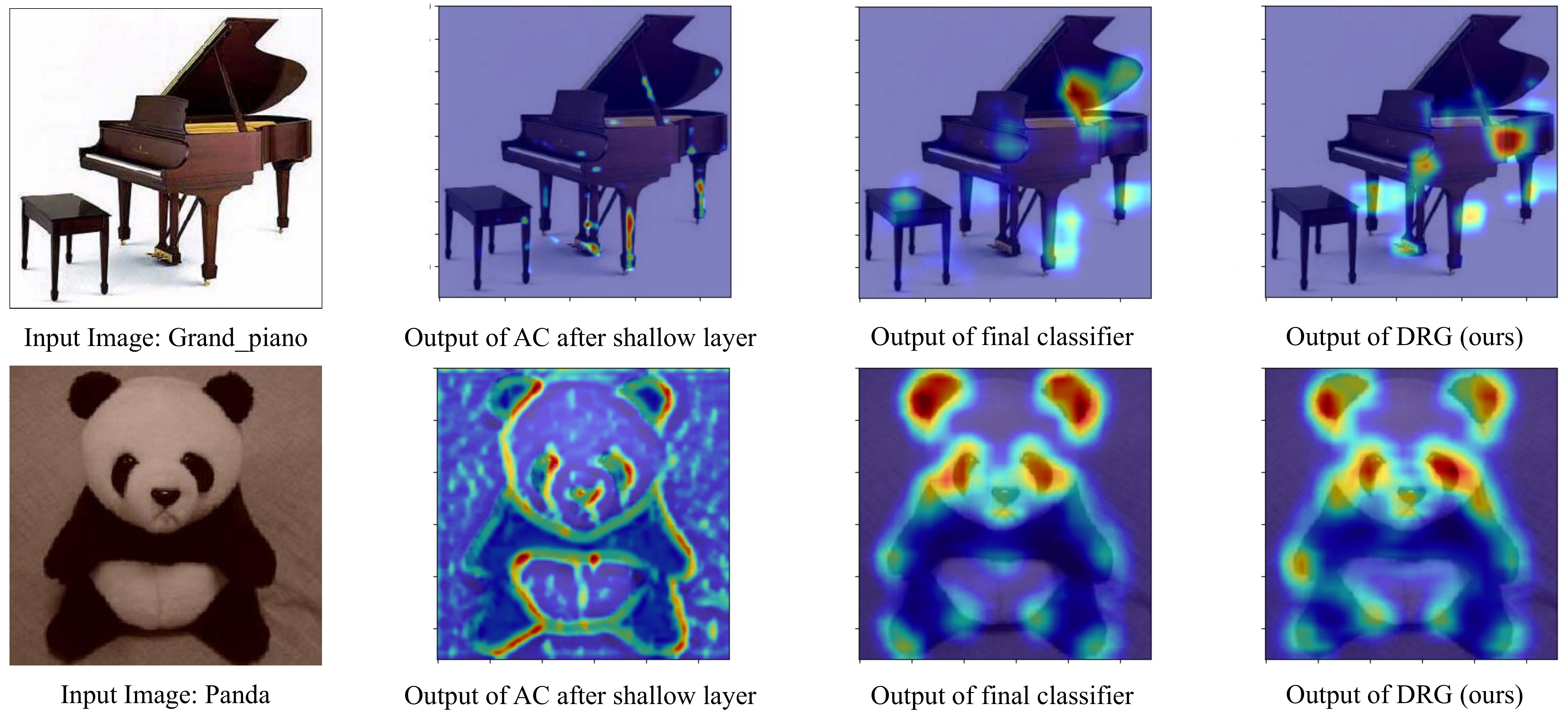}}
\caption{GradCAM heatmaps of different methods on  Caltech101 over ResNet18. From left to right:  input images, output of AC after shallow layer, output of model by BYOT, and the output of DSR (ours). As the heatmaps exemplify, instead of excessive care of one single feature, DRG merges the feature of both  classifiers after the shallow layer and the whole model. }
\label{fig:gradcam}
\end{center}
\end{figure*}

\section{Discussion}
In this section, we discuss the rationality behind our proposed methods through experiments. First, we show the capacity of DRG in information fusion. Then, we analyze the double-effect of DSR in enlarging decision boundary and label smoothing.

\begin{figure*}[ht]
\begin{center}
\centerline{\includegraphics[width=1.80\columnwidth]{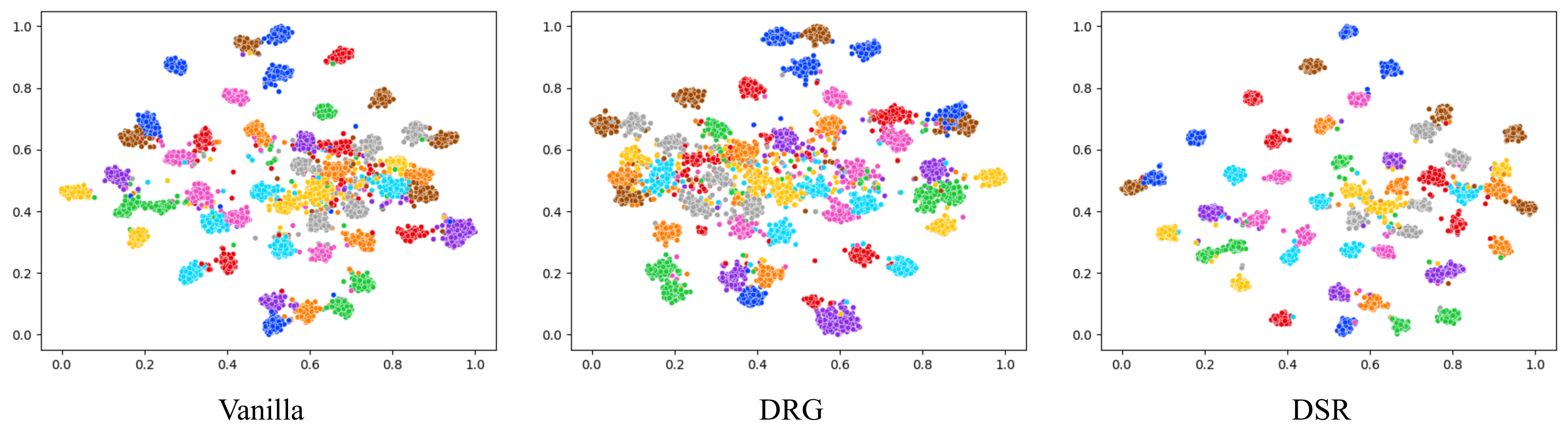}}
\vspace{-5pt}
\caption{FIT-SNE results of  different methods for randomly chosen 50 classes on CIFAR100, ResNet18. DSR 
exerts a more clear decision boundary among classes than the vanilla single model training and DRG.  }
\label{fig:tsne}
\end{center}
\end{figure*}

\subsection{Informulation Fusion of DRG}
DRG achieves the information fusion of features extracted from different parts of a neural network model. To illustrate this, we employ GradCAM \cite{selvaraju2017grad} to virtualize the features characterized by different parts of the model and our DRG method. GradCAM is a method for generating attention heatmaps to visualize the focusing position of a model for the input data. We present the GradCAM results of the output of AC after the shallow layer (i.e., the second block of ResNet18 in our experiments), the output of the whole model, and out DRG method in Fig. \ref{fig:gradcam}.

The results show that the classifier after the shallow layer mainly focuses on  the edge and shape features of the input date, such as the legs of the table and the outline of the panda. In contrast, the whole model with more layers forgets edge features and extracts more determined information, such as the ears of the panda. 
By using the classifier after the shallow layer as the ``poor'' teacher of KD, DRG can capture both edge and detailed information of the input data,  providing valuable insights into the information fusion of our DRG method.

\subsection{Double-effect of DSR}
We can interpret the rationality behind DSR from the following two perspectives.

First, DSR is capable of achieving the consensus of ranked model output probability, which enlarges the decision boundary among different classes. Fig. \ref{fig:tsne} demonstrates the virtualized decision boundary of DRG and DSR over (CIFAR100, ResNet18) using FIT-SNE \cite{linderman2019fast} results \footnote{We use the official implementation of FIT-SNE as \url{https://github.com/KlugerLab/FIt-SNE}}. We randomly sample 50 classes to clearly show the FIT-SNE virtualization. We can observe that our DSR method exerts a more clear decision boundary than vanilla single model training and DRG.

Moreover, DSR is equivalent to a label-smoothing method that progressively designs a label from a distribution rather than a predetermined shape. Specifically, the ``soft'' label used in DSR is the ranked label of another data sample, which is randomly sampled from the dataset. This contributes to a better generalization of DSR. 

 \section{Conclusion}
 In this paper, we propose a lightweight SKD framework with two methods, DRG and DSR, to promote multi-source information fusion and improve the performance of SKD.
 We construct only one auxiliary teacher in DRG and highlight the inter-class model output shape in DSR to achieve better test accuracy with a low time cost. 
Experimental results over enormous datasets and models show that DRG and DSR, and their combination, outperform the baselines with lower or competitive time costs and better robustness. 
In summary, our proposed methods demonstrate significant improvements in self-knowledge distillation through novel approaches to multi-source information fusion.

\bibliographystyle{IEEEtran}
\bibliography{ref}{}

\begin{thebibliography}{10}
\providecommand{\url}[1]{#1}
\csname url@samestyle\endcsname
\providecommand{\newblock}{\relax}
\providecommand{\bibinfo}[2]{#2}
\providecommand{\BIBentrySTDinterwordspacing}{\spaceskip=0pt\relax}
\providecommand{\BIBentryALTinterwordstretchfactor}{4}
\providecommand{\BIBentryALTinterwordspacing}{\spaceskip=\fontdimen2\font plus
\BIBentryALTinterwordstretchfactor\fontdimen3\font minus
  \fontdimen4\font\relax}
\providecommand{\BIBforeignlanguage}[2]{{%
\expandafter\ifx\csname l@#1\endcsname\relax
\typeout{** WARNING: IEEEtran.bst: No hyphenation pattern has been}%
\typeout{** loaded for the language `#1'. Using the pattern for}%
\typeout{** the default language instead.}%
\else
\language=\csname l@#1\endcsname
\fi
#2}}
\providecommand{\BIBdecl}{\relax}
\BIBdecl

\bibitem{hinton2015distilling}
G.~Hinton, O.~Vinyals, and J.~Dean, ``Distilling the knowledge in a neural
  network,'' \emph{arXiv preprint arXiv:1503.02531}, 2015.

\bibitem{furlanello2018born}
T.~Furlanello, Z.~Lipton, M.~Tschannen, L.~Itti, and A.~Anandkumar, ``Born
  again neural networks,'' in \emph{International Conference on Machine
  Learning}.\hskip 1em plus 0.5em minus 0.4em\relax PMLR, 2018, pp. 1607--1616.

\bibitem{zhang2019your}
L.~Zhang, J.~Song, A.~Gao, J.~Chen, C.~Bao, and K.~Ma, ``Be your own teacher:
  Improve the performance of convolutional neural networks via self
  distillation,'' in \emph{Proceedings of the IEEE/CVF International Conference
  on Computer Vision}, 2019, pp. 3713--3722.

\bibitem{ji2021refine}
M.~Ji, S.~Shin, S.~Hwang, G.~Park, and I.-C. Moon, ``Refine myself by teaching
  myself: Feature refinement via self-knowledge distillation,'' in
  \emph{Proceedings of the IEEE/CVF conference on computer vision and pattern
  recognition}, 2021, pp. 10\,664--10\,673.

\bibitem{li2022distilling}
S.~Li, M.~Lin, Y.~Wang, Y.~Wu, Y.~Tian, L.~Shao, and R.~Ji, ``Distilling a
  powerful student model via online knowledge distillation,'' \emph{IEEE
  Transactions on Neural Networks and Learning Systems}, 2022.

\bibitem{phuong2019distillation}
M.~Phuong and C.~H. Lampert, ``Distillation-based training for multi-exit
  architectures,'' in \emph{Proceedings of the IEEE/CVF international
  conference on computer vision}, 2019, pp. 1355--1364.

\bibitem{yang2019snapshot}
C.~Yang, L.~Xie, C.~Su, and A.~L. Yuille, ``Snapshot distillation:
  Teacher-student optimization in one generation,'' in \emph{Proceedings of the
  IEEE/CVF Conference on Computer Vision and Pattern Recognition}, 2019, pp.
  2859--2868.

\bibitem{kim2021self}
K.~Kim, B.~Ji, D.~Yoon, and S.~Hwang, ``Self-knowledge distillation with
  progressive refinement of targets,'' in \emph{Proceedings of the IEEE/CVF
  International Conference on Computer Vision}, 2021, pp. 6567--6576.

\bibitem{shen2022self}
Y.~Shen, L.~Xu, Y.~Yang, Y.~Li, and Y.~Guo, ``Self-distillation from the last
  mini-batch for consistency regularization,'' in \emph{Proceedings of the
  IEEE/CVF Conference on Computer Vision and Pattern Recognition}, 2022, pp.
  11\,943--11\,952.

\bibitem{yuan2020revisiting}
L.~Yuan, F.~E. Tay, G.~Li, T.~Wang, and J.~Feng, ``Revisiting knowledge
  distillation via label smoothing regularization,'' in \emph{Proceedings of
  the IEEE/CVF Conference on Computer Vision and Pattern Recognition}, 2020,
  pp. 3903--3911.

\bibitem{szegedy2016rethinking}
C.~Szegedy, V.~Vanhoucke, S.~Ioffe, J.~Shlens, and Z.~Wojna, ``Rethinking the
  inception architecture for computer vision,'' in \emph{Proceedings of the
  IEEE conference on computer vision and pattern recognition}, 2016, pp.
  2818--2826.

\bibitem{kim2022ai}
H.~Kim, S.~Suh, S.~Baek, D.~Kim, D.~Jeong, H.~Cho, and J.~Kim, ``Ai-kd:
  Adversarial learning and implicit regularization for self-knowledge
  distillation,'' \emph{arXiv preprint arXiv:2211.10938}, 2022.

\bibitem{goodfellow2020generative}
I.~Goodfellow, J.~Pouget-Abadie, M.~Mirza, B.~Xu, D.~Warde-Farley, S.~Ozair,
  A.~Courville, and Y.~Bengio, ``Generative adversarial networks,''
  \emph{Communications of the ACM}, vol.~63, no.~11, pp. 139--144, 2020.

\bibitem{yun2020regularizing}
S.~Yun, J.~Park, K.~Lee, and J.~Shin, ``Regularizing class-wise predictions via
  self-knowledge distillation,'' in \emph{Proceedings of the IEEE/CVF
  conference on computer vision and pattern recognition}, 2020, pp.
  13\,876--13\,885.

\bibitem{xu2019data}
T.-B. Xu and C.-L. Liu, ``Data-distortion guided self-distillation for deep
  neural networks,'' in \emph{Proceedings of the AAAI Conference on Artificial
  Intelligence}, vol.~33, no.~01, 2019, pp. 5565--5572.

\bibitem{kim2018paraphrasing}
J.~Kim, S.~Park, and N.~Kwak, ``Paraphrasing complex network: Network
  compression via factor transfer,'' \emph{Advances in neural information
  processing systems}, vol.~31, 2018.

\bibitem{li2022curriculum}
Z.~Li, X.~Li, L.~Yang, B.~Zhao, R.~Song, L.~Luo, J.~Li, and J.~Yang,
  ``Curriculum temperature for knowledge distillation,'' \emph{arXiv preprint
  arXiv:2211.16231}, 2022.

\bibitem{li2022asymmetric}
X.-C. Li, W.-S. Fan, S.~Song, Y.~Li, B.~Li, Y.~Shao, and D.-C. Zhan,
  ``Asymmetric temperature scaling makes larger networks teach well again,''
  \emph{arXiv preprint arXiv:2210.04427}, 2022.

\bibitem{romero2014fitnets}
A.~Romero, N.~Ballas, S.~E. Kahou, A.~Chassang, C.~Gatta, and Y.~Bengio,
  ``Fitnets: Hints for thin deep nets,'' \emph{arXiv preprint arXiv:1412.6550},
  2014.

\bibitem{heo2019comprehensive}
B.~Heo, J.~Kim, S.~Yun, H.~Park, N.~Kwak, and J.~Y. Choi, ``A comprehensive
  overhaul of feature distillation,'' in \emph{Proceedings of the IEEE/CVF
  International Conference on Computer Vision}, 2019, pp. 1921--1930.

\bibitem{chen2019data}
H.~Chen, Y.~Wang, C.~Xu, Z.~Yang, C.~Liu, B.~Shi, C.~Xu, C.~Xu, and Q.~Tian,
  ``Data-free learning of student networks,'' in \emph{Proceedings of the
  IEEE/CVF International Conference on Computer Vision}, 2019, pp. 3514--3522.

\bibitem{binici2022robust}
K.~Binici, S.~Aggarwal, N.~T. Pham, K.~Leman, and T.~Mitra, ``Robust and
  resource-efficient data-free knowledge distillation by generative pseudo
  replay,'' in \emph{Proceedings of the AAAI Conference on Artificial
  Intelligence}, vol.~36, no.~6, 2022, pp. 6089--6096.

\bibitem{zhao2022decoupled}
B.~Zhao, Q.~Cui, R.~Song, Y.~Qiu, and J.~Liang, ``Decoupled knowledge
  distillation,'' in \emph{Proceedings of the IEEE/CVF Conference on computer
  vision and pattern recognition}, 2022, pp. 11\,953--11\,962.

\bibitem{wang2019private}
J.~Wang, W.~Bao, L.~Sun, X.~Zhu, B.~Cao, and S.~Y. Philip, ``Private model
  compression via knowledge distillation,'' in \emph{Proceedings of the AAAI
  Conference on Artificial Intelligence}, vol.~33, no.~01, 2019, pp.
  1190--1197.

\bibitem{ma2021undistillable}
H.~Ma, T.~Chen, T.-K. Hu, C.~You, X.~Xie, and Z.~Wang, ``Undistillable: Making
  a nasty teacher that cannot teach students,'' \emph{arXiv preprint
  arXiv:2105.07381}, 2021.

\bibitem{kundu2021analyzing}
S.~Kundu, Q.~Sun, Y.~Fu, M.~Pedram, and P.~Beerel, ``Analyzing the
  confidentiality of undistillable teachers in knowledge distillation,''
  \emph{Advances in Neural Information Processing Systems}, vol.~34, pp.
  9181--9192, 2021.

\bibitem{jandial2022distilling}
S.~Jandial, Y.~Khasbage, A.~Pal, V.~N. Balasubramanian, and B.~Krishnamurthy,
  ``Distilling the undistillable: Learning from a nasty teacher,'' in
  \emph{Computer Vision--ECCV 2022: 17th European Conference, Tel Aviv, Israel,
  October 23--27, 2022, Proceedings, Part XIII}.\hskip 1em plus 0.5em minus
  0.4em\relax Springer, 2022, pp. 587--603.

\bibitem{zhang2021delving}
C.-B. Zhang, P.-T. Jiang, Q.~Hou, Y.~Wei, Q.~Han, Z.~Li, and M.-M. Cheng,
  ``Delving deep into label smoothing,'' \emph{IEEE Transactions on Image
  Processing}, vol.~30, pp. 5984--5996, 2021.

\bibitem{liang2022efficient}
J.~Liang, L.~Li, Z.~Bing, B.~Zhao, Y.~Tang, B.~Lin, and H.~Fan, ``Efficient one
  pass self-distillation with zipf’s label smoothing,'' in \emph{Computer
  Vision--ECCV 2022: 17th European Conference, Tel Aviv, Israel, October
  23--27, 2022, Proceedings, Part XI}.\hskip 1em plus 0.5em minus 0.4em\relax
  Springer, 2022, pp. 104--119.

\bibitem{he2016deep}
K.~He, X.~Zhang, S.~Ren, and J.~Sun, ``Deep residual learning for image
  recognition,'' in \emph{Proceedings of the IEEE conference on computer vision
  and pattern recognition}, 2016, pp. 770--778.

\bibitem{krizhevsky2009learning}
A.~Krizhevsky, G.~Hinton \emph{et~al.}, ``Learning multiple layers of features
  from tiny images,'' 2009.

\bibitem{deng2009imagenet}
J.~Deng, W.~Dong, R.~Socher, L.-J. Li, K.~Li, and L.~Fei-Fei, ``Imagenet: A
  large-scale hierarchical image database,'' in \emph{2009 IEEE conference on
  computer vision and pattern recognition}.\hskip 1em plus 0.5em minus
  0.4em\relax Ieee, 2009, pp. 248--255.

\bibitem{xie2017aggregated}
S.~Xie, R.~Girshick, P.~Doll{\'a}r, Z.~Tu, and K.~He, ``Aggregated residual
  transformations for deep neural networks,'' in \emph{Proceedings of the IEEE
  conference on computer vision and pattern recognition}, 2017, pp. 1492--1500.

\bibitem{huang2017densely}
G.~Huang, Z.~Liu, L.~Van Der~Maaten, and K.~Q. Weinberger, ``Densely connected
  convolutional networks,'' in \emph{Proceedings of the IEEE conference on
  computer vision and pattern recognition}, 2017, pp. 4700--4708.

\bibitem{devries2017improved}
T.~DeVries and G.~W. Taylor, ``Improved regularization of convolutional neural
  networks with cutout,'' \emph{arXiv preprint arXiv:1708.04552}, 2017.

\bibitem{selvaraju2017grad}
R.~R. Selvaraju, M.~Cogswell, A.~Das, R.~Vedantam, D.~Parikh, and D.~Batra,
  ``Grad-cam: Visual explanations from deep networks via gradient-based
  localization,'' in \emph{Proceedings of the IEEE international conference on
  computer vision}, 2017, pp. 618--626.

\bibitem{linderman2019fast}
G.~C. Linderman, M.~Rachh, J.~G. Hoskins, S.~Steinerberger, and Y.~Kluger,
  ``Fast interpolation-based t-sne for improved visualization of single-cell
  rna-seq data,'' \emph{Nature methods}, vol.~16, no.~3, pp. 243--245, 2019.

\end{thebibliography}

\vspace{11pt}

\vspace{11pt}

\vfill

\end{document}